\documentclass[11pt]{article}

\usepackage[final]{acl}

\usepackage{times}
\usepackage{latexsym}

\usepackage[T1]{fontenc}
\usepackage[utf8]{inputenc}
\usepackage{microtype}
\usepackage{inconsolata}

\usepackage{graphicx}
\usepackage{booktabs}
\usepackage{multirow}
\usepackage{amsmath}
\usepackage{amssymb}
\usepackage{tikz}
\usetikzlibrary{arrows.meta, positioning, calc, fit, backgrounds}

\definecolor{pdense}{HTML}{2E6E9E}   
\definecolor{pkg}{HTML}{B7791F}      
\definecolor{presolve}{HTML}{2F855A} 
\definecolor{pread}{HTML}{6B46C1}    
\definecolor{pbg}{HTML}{F5F7FA}      

\title{SPARQL-Planned Single-Hop Traversal over a\\ Live Knowledge Graph for Multi-Hop Question Answering}

\author{
  Priyesh Vakharia \quad Masud Moshtaghi \quad Pratyay Banerjee \\
  \textbf{Nicholas Jedema} \quad \textbf{Amita Misra} \quad \textbf{Ankit Chadha} \\
  Amazon \\
  \texttt{\{priyvak, mmasud, pratyay, jedem, misrami, ankitrc\}@amazon.com}
}

\begin{document}
\maketitle

\begin{abstract}
Semantic retrieval struggles with multi-hop question answering on both recall and precision: gold passages are buried among distractors that degrade the reader's answer, and retrieving them requires reasoning over the query rather than matching it. GraphRAG techniques expand semantic retrieval by connecting passages to graph nodes and performing one or more hops of expansion after initial retrieval. This improves recall, but multi-hop expansion becomes expensive and brings additional noise into the retrieved set. We break this tradeoff by \emph{interleaving} dense retrieval with knowledge-graph traversal for complex multi-hop questions. Dense retrieval locates the first relevant regions of the search space at good recall, while traversal of a pre-existing external KG (Wikidata) restores precision and improves recall by confirming retrieved passages against a reasoning path through the KG. To keep traversal both question-adaptive and reliable, we have an LLM plan it as code generation task for graph traversal in SPARQL format but \emph{parse rather than execute} the query. On full 1{,}000-question splits under a shared corpus and reader, and with a single configuration applied identically to every dataset, we reach \textbf{62.5 / 75.1} EM/F1 on HotpotQA, \textbf{73.5 / 81.0} on 2WikiMultiHopQA, and \textbf{37.1 / 50.0} on MuSiQue, ahead of reproduced HippoRAG2, T\textsuperscript{2}RAG, and IRCoT on all three benchmarks. A second, token-matched setting stays ahead of T\textsuperscript{2}RAG on HotpotQA and 2Wiki while spending roughly its per-query budget. Executing the generated SPARQL instead of parsing it costs up to 11.4 EM, mostly because raw queries return empty results, not because of syntax errors.
\end{abstract}

\section{Introduction}

Retrieval-Augmented Generation (RAG) \citep{lewis2021retrievalaugmentedgenerationknowledgeintensivenlp} grounds large language models (LLMs) in factual, up-to-date knowledge by supplementing parametric memory with retrieved documents \citep{karpukhin2020dpr} \citep{gao2023retrieval}. Standard RAG treats retrieval as a function of the query alone: the system encodes the question, searches a passage index, and returns the top-$k$ results by vector similarity. For simple factoid questions this suffices, but for multi-hop reasoning, vector similarity over the original query misses documents that are essential yet semantically distant from the surface question. These \emph{bridging} documents can be discovered by reasoning about entity relationships.

A natural fix is to widen the net, but dense retrieval buys recall at a steep precision cost: on HotpotQA \citep{yang2018hotpotqa} the top-10 dense passages reach 80\% recall at only 32\% precision (Section~\ref{sec:results}), forcing the reader to reason over a context that is mostly distractors. Raising $k$ dilutes precision further; lowering it drops gold.

Graph-augmented retrieval (GraphRAG) \citep{peng2024graphretrievalaugmentedgenerationsurvey} pursues this structure, but the dominant paradigm builds a graph \emph{from the target corpus}: an LLM extracts entity–relation triplets from every passage \citep{edge2024graphrag,guo2024lightrag}. These triplets are fragmented and noisy, limiting the multi-hop reasoning they are meant to support \citep{kim2026thrag}, and the extraction is expensive construction cost \citep{li2026dualgraphrag}, which make the approach more suitable for closed-domain,i.e., finance, sport. We instead traverse a \emph{pre-existing external} KG, Wikidata, whose structure is already dense and extraction-free. Prior external-KG methods, however, traverse it with fixed, query-independent heuristics \citep{wu2024kginfused,zhu2024kg2rag} rather than adapting the traversal to each question.

Our key idea is to \emph{interleave} the two retrievers in a query-refinement loop so each supplies what the other lacks. Dense retrieval seeds entities and keeps recall high, and graph traversal confirms which seeds lie on the reasoning path and fetches the bridging passages dense retrieval missed, restoring precision (on HotpotQA, from 32\% to 81\% precision at better recall). This interleaving, rather than either component alone, breaks the tradeoff.

On the full 1{,}000-question evaluation splits of three multi-hop benchmarks, with a shared corpus, retriever, and reader across all systems, our approach outperforms strong reproduced baselines on all three, with the largest margins where Wikidata coverage is strongest. Our contributions are:

\begin{itemize}
  \item \textbf{Interleaving dense retrieval with KG traversal} through a batch resolution and query-refinement loop, so dense recall and KG precision reinforce each other.
  \item \textbf{Parse-don't-execute traversal}, the enabling mechanism: the LLM plans traversal by writing SPARQL, but rather than run the query, which mostly returns empty results on mislinked identifiers, label mismatches, or over-constrained patterns (syntax errors and timeouts are rare), we parse it into tuples and execute anchored single-hop lookups. A controlled ablation shows that executing the raw query instead costs up to 11.4 EM (\S\ref{sec:parse_vs_execute}).
  \item \textbf{A controlled $n{=}1000$ study} against reproduced HippoRAG2, T\textsuperscript{2}RAG, and IRCoT under an identical corpus/retriever/reader/LLM, with component ablations and a cost--accuracy frontier: matched to T\textsuperscript{2}RAG's per-query token budget we remain more accurate on HotpotQA and 2Wiki.
  \item \textbf{Two operating points along a cost--accuracy frontier}, both applied identically to every dataset: an \emph{unbounded} setting that pairs a lower resolver bar with a full, commit-oriented reader context for maximum accuracy, and a \emph{bounded} setting whose token-reduction levers match T\textsuperscript{2}RAG's per-query budget while remaining more accurate than T\textsuperscript{2}RAG on HotpotQA and 2Wiki.
\end{itemize}

We use LLM-generated SPARQL as the planning medium because LLMs reason and plan more reliably through structured outputs such as code. SPARQL and Wikidata schemas are well represented in typical pretraining corpora, so the model already knows how to express a textual query as a well-formed graph pattern. Building on this intuition, our method improves both answer quality and retrieval quality across all three datasets relative to the baselines, reaching 62.5, 73.5, and 37.1 EM on HotpotQA, 2WikiMultiHopQA, and MuSiQue---ahead of HippoRAG2, T\textsuperscript{2}RAG, and IRCoT on every split.
\section{Related Work}

\paragraph{Corpus-constructed graphs.} Microsoft's GraphRAG \citep{edge2024graphrag} builds a graph from the target corpus by extracting entity–relation triplets with an LLM, then clusters them into communities and generates hierarchical summaries. LightRAG \citep{guo2024lightrag} similarly constructs an entity graph and performs dual-level retrieval, while RAPTOR \citep{sarthi2024raptor} recursively clusters and summarizes passages into a retrieval tree and KGP \citep{wang2024knowledge} builds a document graph an LLM agent then traverses. Because LLM-generated triplets are fragmented and sparsely connected, TH-RAG \citep{kim2026thrag} organizes them into a topic hierarchy to improve connectivity, and DualGraphRAG \citep{li2026dualgraphrag} retrieves both local triples and shortest paths while cutting construction time by an order of magnitude. All of these require an expensive per-corpus build and apply uniform traversal regardless of the query.

\paragraph{Iterative retrieval for multi-hop QA.} A separate line interleaves retrieval with reasoning without any graph. MDR \citep{xiong2021mdr} performs iterative dense retrieval, re-encoding the query with each retrieved passage to fetch the next hop; IRCoT \citep{trivedi2023interleavingretrievalchainofthoughtreasoning} alternates chain-of-thought steps with retrieval, Self-Ask \citep{press2023measuring} decomposes a question into follow-up sub-questions, and FLARE \citep{jiang2023active} triggers retrieval whenever the generator is uncertain. Our query-refinement loop is in this spirit, but the rewritten query is grounded in structured facts recovered from the KG rather than generated free-form, and it drives graph traversal as well as dense retrieval. On the KG side, GNN-RAG \citep{mavromatis2024gnnrag} instead trains a graph neural network to retrieve reasoning subgraphs, whereas our traversal is training-free.

\paragraph{Agentic and reasoning-augmented GraphRAG.} A parallel line integrates LLM reasoning more tightly with structure. Youtu-GraphRAG \citep{dong2026youtugraphrag} adds hierarchical community detection in an agentic framework; \citet{zarrinkia2026sparql} use SPARQL-based chain-of-thought prompting with graph-walk compression. T\textsuperscript{2}RAG \citep{gong2026t2rag} avoids explicit traversal, decomposing questions into search triples resolved iteratively against a proposition index. These improve accuracy but pay multiple LLM calls per query for traversal or triple resolution.

\paragraph{External-KG augmentation and traversal reasoning.} KG-Infused RAG \citep{wu2024kginfused} and KG\textsuperscript{2}RAG \citep{zhu2024kg2rag} expand retrieval using a pre-existing KG such as Wikidata, avoiding extraction cost and noise, but apply fixed-radius or fixed-parameter spreading activation, and KAPING \citep{baek2023knowledge} simply prepends retrieved triples to the prompt without traversal. Think-on-Graph \citep{sun2024thinkongraph} and its text-augmented successor ToG-2 \citep{ma2024think} treat the KG as an environment for LLM-guided beam search; RoG \citep{luo2024reasoning} first plans relation paths on the KG and then retrieves along them; StructGPT \citep{jiang2023structgpt} gives the LLM tool calls to read structured data iteratively; GraphFlow \citep{yu2025graphflow} trains a GFlowNet for trajectory sampling; ARK \citep{polonuer2026ark} is a tool-using agent balancing breadth and depth. These either apply uniform budgets, require graph-specific training, or incur full LLM inference at every traversal step. We share their use of a pre-existing KG but differ in \emph{how} we plan the traversal. ToG and RoG already achieve question-specific, execution-failure-free traversal, but at a cost: ToG invokes the LLM at every hop to pick the next relation, and RoG fine-tunes a dedicated relation-path planner. Methods that instead have an LLM emit executable SPARQL inherit its failure modes (hallucinated identifiers, syntax errors, timeouts, empty results); the closest, the concurrent MARS \citep{srivastava2026mars}, likewise performs adaptive multi-hop KGQA without fine-tuning but \emph{generates and executes} SPARQL and so remains exposed to them. Our distinction is to treat generated SPARQL as a \emph{plan rather than an executable artifact}: a single planning call emits SPARQL that we parse into anchored single-hop lookups, interleaved with dense retrieval, with no per-step LLM inference, no execution risk, and no traversal-specific training. We are thus not the first to reach failure-free question-specific traversal, but the first to reach it by parsing rather than executing generated SPARQL over a live external KG.

\section{Method}

\subsection{Task Formulation}
\label{sec:task}

Given a multi-hop question $q$, a corpus of passages $P=\{p_1,\dots,p_n\}$ each with a title $t_i$, and a knowledge graph $K$ (Wikidata) whose entities correspond to passage titles, the goal is to retrieve the gold supporting passages $G_q \subset P$ and generate an answer from their text. Benchmarks annotate each question with 2–4 gold passages whose titles are entities on the reasoning chain. The reader generates the final answer conditioned \emph{only} on passage text. The KG is used purely for retrieval guidance, so that answers remain grounded in natural-language evidence rather than raw triples.

\subsection{Overview}

\begin{figure*}[t]
  \centering
  \begin{tikzpicture}[
      font=\small,
      >={Stealth[length=2mm]},
      box/.style={rounded corners=3pt, draw, very thick, align=center,
                  minimum height=11mm, inner sep=5pt},
      io/.style={align=center, font=\small\bfseries},
      lbl/.style={font=\scriptsize, align=center, inner sep=1.5pt, fill=white},
    ]
    \node[box, draw=presolve, fill=presolve!8, text width=32mm] (res)
      {\textbf{Resolve \& refine}\\[1pt]\scriptsize confirm on-path passages,\\
       extract facts, rewrite query};
    \node[box, draw=pdense, fill=pdense!8, text width=26mm,
          below left=15mm and 4mm of res] (dpr)
      {\textbf{Dense retrieval}\\[1pt]\scriptsize seed \& re-seed\\ candidate passages};
    \node[box, draw=pkg, fill=pkg!14, text width=34mm,
          below right=15mm and 4mm of res] (plan)
      {\textbf{Plan (KG traversal)}\\[1pt]\scriptsize write SPARQL $\rightarrow$
       \emph{parse, not execute}\\ $\rightarrow$ single-hop Wikidata lookups};

    \node[io, left=12mm of dpr] (q) {Question};
    \node[box, draw=pread, fill=pread!8, text width=13mm, right=13mm of res] (read)
      {\textbf{Reader}};
    \node[io, right=8mm of read] (ans) {Answer};

    \draw[->, thick] (q) -- (dpr);

    \draw[->, pdense, thick] (dpr) -- (res)
      node[lbl, midway, sloped]{seed passages};
    \draw[->, presolve, thick] (res) -- (plan)
      node[lbl, midway, sloped]{still needed};
    \draw[->, pkg, thick] (plan) -- (dpr)
      node[lbl, midway, below]{bridge entities, refined query};

    \draw[->, pread, thick] (res) -- (read)
      node[lbl, midway, above]{answerable};
    \draw[->, thick] (read) -- (ans);
  \end{tikzpicture}
  \caption{The interleaved pipeline as a three-agent loop.
  \textcolor{pdense}{\textbf{Dense retrieval}} seeds candidate passages from the
  question; the \textcolor{presolve}{\textbf{resolver}} confirms which lie on the
  reasoning path, extracts facts, and states what is still needed; the
  \textcolor{pkg}{\textbf{plan agent}} turns that into SPARQL which we
  \emph{parse, not execute} into anchored single-hop Wikidata lookups, whose
  discovered bridge entities sharpen a refined query that re-seeds dense retrieval.
  The loop repeats until the resolver judges the question answerable, after which
  the \textcolor{pread}{\textbf{reader}} answers from the confirmed passages.}
  \label{fig:pipeline}
\end{figure*}
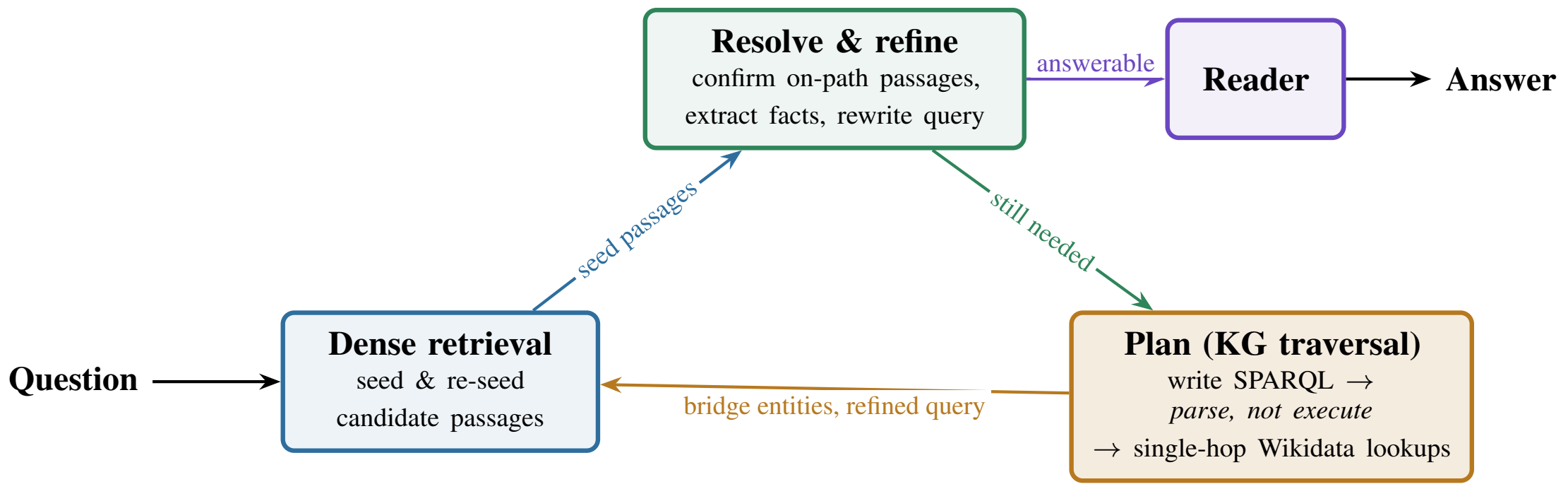

Figure~\ref{fig:pipeline} shows the pipeline as a three-agent loop. Dense retrieval seeds entities from the question and keeps recall high. A batch resolution agent reads the seed passages, extracts the facts on the reasoning path, and rewrites the question with resolved entities filled in. A plan agent expresses the next traversal as SPARQL, which we parse (rather than execute) into single-hop lookups against Wikidata; discovered bridge entities are title-matched to corpus passages and routed back through the resolver, which either confirms passages or emits a sharper refined query that re-seeds dense retrieval. The loop repeats until the resolver judges the question answerable, no progress is made, or a maximum-iteration budget is reached, after which a reader answers from the confirmed passages. Dense retrieval and KG traversal thus feed each other: retrieved text supplies the named entities the plan agent anchors on, and KG facts sharpen the query that dense retrieval acts on.

\subsection{SPARQL as a Traversal Plan}
\label{sec:parse}

The core mechanism separates traversal \emph{reasoning} from \emph{execution}. A \textbf{plan agent} is prompted with the question, the facts resolved so far, the information still needed, the properties already confirmed to exist on known entities, and a list of dead ends. It emits one or more SPARQL queries. Rather than executing them, we parse each query into a set of traversal steps
\[
(e_i,\; r_i,\; d_i), \quad d_i \in \{\text{out}, \text{in}\},
\]
where $e_i$ is an entity label, $r_i$ a Wikidata property, and $d_i$ the traversal direction. The parser recognizes label anchors (\texttt{?x rdfs:label "Name"@en}), direct QID anchors (\texttt{wd:Q42 wdt:P69 ?u}), and incoming patterns (\texttt{?w wdt:P57 wd:Q25191}); it discards type constraints (\texttt{P31}/\texttt{P279}) and \texttt{SERVICE}/\texttt{FILTER} blocks, detects and drops placeholder QIDs the model invents when it does not know a real ID, and chains dependent hops via a \texttt{\$RESULT} reference when one hop's output feeds the next.

Each extracted step is executed as an anchored single-hop query, \texttt{wd:QID wdt:$r_i$ ?v} or its reverse, preceded by resolving $e_i$ to a QID through the entity-search API. Because each query binds a concrete entity and contains a single triple pattern, it returns instantly and never times out. This buys the LLM's reasoning about \emph{which} relationships matter while making execution robust: syntax errors, timeouts, and query-complexity failures are structurally impossible. After each traversal we \emph{probe} the discovered entities for their available properties and feed this list back to the plan agent, which corrects property-selection errors (e.g., confusing \texttt{P570} ``date of death'' with \texttt{P20} ``place of death'').

\begin{table*}[t]
  \centering
  \begin{tabular}{lcccccc}
    \toprule
    & \multicolumn{2}{c}{\textbf{HotpotQA}} & \multicolumn{2}{c}{\textbf{2WikiMultiHopQA}} & \multicolumn{2}{c}{\textbf{MuSiQue}} \\
    \cmidrule(lr){2-3}\cmidrule(lr){4-5}\cmidrule(lr){6-7}
    \textbf{Method} & EM & F1 & EM & F1 & EM & F1 \\
    \midrule
    DPR-only\textsuperscript{$\dagger$} & 48.0 & 58.3 & 32.0 & 34.9 & 13.0 & 18.8 \\
    LightRAG\textsuperscript{$\S$} & 0.0 & 1.3 & 0.0 & 5.4 & 0.0 & 0.8 \\
    HippoRAG2            & 53.5 & 66.6 & 35.2 & 40.8 & \underline{28.6} & \underline{40.4} \\
    T\textsuperscript{2}RAG & 48.4 & 61.0 & 55.2 & 63.3 & 26.6 & 37.7 \\
    IRCoT               & \underline{57.7} & \underline{70.5} & 53.1 & 60.5 & 26.4 & 38.8 \\
    \midrule
    Ours (bounded)\textsuperscript{$\ddagger$} & 57.2 & 69.8 & \underline{67.7} & \underline{75.1} & 27.3 & 36.1 \\
    Ours (unbounded)        & \textbf{62.5} & \textbf{75.1} & \textbf{73.5} & \textbf{81.0} & \textbf{37.1} & \textbf{50.0} \\
    
    \bottomrule
  \end{tabular}
  \caption{Answer quality (EM / F1) on full 1{,}000-question splits. Shared corpus, retriever, reader, and LLM (Qwen3-32B). We report our method at two operating points, applied \emph{identically} to all three datasets: \emph{Ours (unbounded)} uses a lower resolver relevance bar with a full commit-oriented reader context, while \emph{Ours (bounded)}\textsuperscript{$\ddagger$} applies the token-reduction levers of \S\ref{sec:efficiency} tuned to T\textsuperscript{2}RAG's per-query budget ($\approx$3.2K tokens). Neither setting is tuned per dataset.\textsuperscript{$\S$}LightRAG emits verbose multi-paragraph answers by design (it targets LLM-judge evaluation, not exact match), which score near-zero under EM even when factually correct. We report its native output for consistency with prior work \citep{gutierrez2025hipporag2,kim2026thrag}; its retrieval quality (Table~\ref{tab:retrieval}) is the format-independent comparison. There is a 3\% margin of error for this sample size for the reported numbers on the benchmarks}
  \label{tab:main}
\end{table*}

\subsection{Batch Resolution and Query Refinement}
\label{sec:resolve}

The \textbf{resolution agent} is the bridge between text and graph. Given the question, the facts resolved so far, and a set of candidate passages, it decides in a \emph{single LLM call} which passages lie on the reasoning path, extracts a fact from each, states what is still needed, and rewrites the question with all resolved entities filled in.

The refined query is the engine of the feedback loop. When KG traversal discovers a bridge entity (e.g., that a painter died in Venice), the resolver folds it into a sharper query that dense retrieval can act on, surfacing the specific passage that the original query could not distinguish from same-titled distractors. Symmetrically, retrieved passages supply the named entities the plan agent anchors its next SPARQL on.

\textbf{Unbounded vs.\ bounded setting}. We expose a \emph{relevance bar} on the resolver as a knob along a cost--accuracy frontier, giving two operating points that we apply identically to every dataset. The \textbf{bounded} setting confirms only passages on the direct reasoning path, maximizing precision at low token cost. The \textbf{unbounded} setting makes two changes: (i) the resolver confirms any passage that helps with the final answer \emph{or any intermediate hop}, trading precision for recall, and (ii) similar to HippoRAG2 \citep{gutierrez2025hipporag2}, the reader is always given a full set of reader-top-$k$ passages (confirmed first, backfilled by reranking the seen pool) under a commit prompt that requires a best-guess answer rather than abstaining. Deep multi-hop questions leave the bounded resolver with too few confirmed passages and lead the reader to refuse when a hop is only partially supported; the unbounded setting supplies more complete evidence and converts those abstentions into answers, at higher per-query cost (Section~\ref{sec:analysis}).

\subsection{Seeding, KG Title-Matching, and $n$-gram Matching}

Three deterministic mechanisms feed the resolver. (i) \textbf{Dense seeding}: Contriever-msmarco retrieves the top-$k$ passages for the (refined) query. (ii) \textbf{KG title-matching}: when SPARQL traverses an entity $X$, $X$ and any discovered bridge entity are matched by normalized title against the corpus, recovering gold passages that dense retrieval never surfaced. (iii) \textbf{Question $n$-gram matching}: contiguous 5$\to$2-word $n$-grams of the question are matched against corpus titles, catching exact entity names that dense retrieval misses (``Billy Elliot'', ``B Boy'') at zero LLM cost. All three route through the resolver, so its relevance filter preserves precision.

\subsection{Answer Generation}

The reader receives the top-$k$ confirmed passages (reranked by the refined query) plus the resolved facts, and produces a concise answer via chain-of-thought prompting. If the resolver already emitted a complete answer at its stop decision, that answer is used directly.

\section{Experimental Setup}

\paragraph{Datasets.} We evaluate on three multi-hop QA benchmarks: HotpotQA \citep{yang2018hotpotqa} (9{,}811-passage corpus), 2WikiMultiHopQA \citep{ho2020wikimultihop} (6{,}119 passages), and MuSiQue \citep{trivedi2022musique} (11{,}656 passages). For each we use the 1{,}000-question evaluation split and its retrieval corpus released by HippoRAG \citep{gutierrez2024hipporag} and subsequently adopted by HippoRAG2 \citep{gutierrez2025hipporag2} and T\textsuperscript{2}RAG \citep{gong2026t2rag}; using the same splits and corpora keeps our numbers directly comparable to that line of work. HotpotQA is predominantly 2-hop; 2Wiki mixes comparison and compositional 2-hop questions; MuSiQue contains 2–4-hop compositional questions and is the hardest.

\paragraph{Baselines.} Along with LightRAG, we reproduce three other strong GraphRAG systems, HippoRAG2 \citep{gutierrez2025hipporag2} and T\textsuperscript{2}RAG \citep{gong2026t2rag}, and the iterative retrieve-and-reason baseline IRCoT \citep{trivedi2023interleavingretrievalchainofthoughtreasoning}, under identical conditions (same corpus, retriever, reader LLM, and question set), so that comparisons are controlled rather than cross-paper. IRCoT is the canonical graph-free multi-hop baseline in this protocol: it interleaves chain-of-thought steps with dense retrieval but applies no confirmation filter, which lets us isolate the contribution of KG-guided confirmation. We stress that these are \emph{matched-harness} reproductions, not the baselines' published configurations: HippoRAG2's leading numbers use NV-Embed-v2 embeddings with a GPT-4o-mini reader, whereas here every system shares Contriever-msmarco and Qwen3-32B to enable a fair comparison between systems using the same backbone. This design isolates the traversal method from retriever and reader strength; a portion of the 2Wiki gap in particular is an abstention effect rather than retrieval quality, which we quantify in Section~\ref{sec:analysis}.

\paragraph{Implementation.} All systems use Qwen3-32B \citep{yang2025qwen3technicalreport} as the LLM and Contriever-msmarco \citep{izacard2022unsuperviseddenseinformationretrieval} with a FAISS flat index as the dense retriever. We report our method at two operating points, each applied \emph{identically} to all three datasets: \emph{Ours (unbounded)} uses \texttt{max\_iterations}$=3$, temperature $0.0$, DPR top-$k=10$, reader top-$k=5$, the lower resolver relevance bar, and a commit-oriented reader with the full passage context; \emph{Ours (bounded)} adds the token-reduction levers of \S\ref{sec:efficiency} (slim schema/plan/resolve prompts, early-stop on the initial dense pool, DPR top-$k{=}5$, reader top-$k{=}3$, \texttt{max\_iterations}$=1$, direct-path-only resolver) to reach roughly T\textsuperscript{2}RAG's per-query token budget.

\paragraph{Metrics.} We report answer Exact Match (EM) and token-level F1, and, for retrieval, supporting-passage recall, precision, and F1.

\begin{table*}[t]
  \centering
  \begin{tabular}{lccccccccc}
    \toprule
    & \multicolumn{3}{c}{\textbf{HotpotQA}} & \multicolumn{3}{c}{\textbf{2WikiMultiHopQA}} & \multicolumn{3}{c}{\textbf{MuSiQue}} \\
    \cmidrule(lr){2-4}\cmidrule(lr){5-7}\cmidrule(lr){8-10}
    \textbf{Method} & R & P & F1 & R & P & F1 & R & P & F1 \\
    \midrule
    DPR-only             & 80.3 & 32.1 & 45.9 & 66.5 & 31.0 & 41.5 & 57.0 & 34.1 & 41.1 \\
    LightRAG             & 43.2 & 5.4 & 9.4 & 67.9 & 10.8 & 18.1 & 34.8 & 4.8 & 8.2 \\
    HippoRAG2            & 86.2 & -- & -- & 66.7 & -- & -- & 60.1 & -- & -- \\
    T\textsuperscript{2}RAG & 79.8 & 31.7 & 45.1 & 82.0 & 35.0 & 48.0 & 63.6 & 32.5 & 41.3 \\
    IRCoT\textsuperscript{$\ast$} & \underline{89.9} & 21.7 & 35.0 & \underline{83.8} & 22.9 & 35.4 & \underline{65.9} & 18.3 & 28.2 \\
    \midrule
    Ours (bounded)       & 79.8 & \textbf{85.8} & \underline{80.9} & \underline{87.2} & \textbf{91.4} & \underline{88.0} & 48.0 & \textbf{63.7} & \underline{52.2} \\
    \textbf{Ours (unbounded)}        & \textbf{91.6} & \underline{80.7} & \textbf{83.6} & \textbf{96.4} & \underline{91.3} & \textbf{92.4} & \textbf{68.0} & \underline{55.6} & \textbf{57.6} \\
    
    \bottomrule
  \end{tabular}
  \caption{Supporting-passage retrieval quality (recall / precision / F1) at $n{=}1000$. KG traversal converts dense retrieval's low-precision pool into a high-precision retrieval set at better recall (a $\sim$2.5--3$\times$ precision gain on HotpotQA and 2Wiki). \emph{Ours (bounded)} preserves this high-precision retrieval at T\textsuperscript{2}RAG's token budget on HotpotQA and 2Wiki (2Wiki precision even rises to 91.4); on MuSiQue the tighter budget lowers recall to 48.0. HippoRAG2 uses top-$k$ retrieval and reports recall@5 only (no precision).\textsuperscript{$\ast$}IRCoT accumulates passages over its interleaved retrieve-and-reason steps. Its high recall comes at very low precision, since it retrieves a fresh top-$k$ per reasoning step without a confirmation filter.}
  \label{tab:retrieval}
\end{table*}

\section{Results}
\label{sec:results}

\subsection{Main Results}

Our system is the best on all three splits.

Under the \emph{unbounded} setting, applied identically to all three datasets, our method reaches 62.5 EM on HotpotQA (exceeding HippoRAG2 by $+9.0$, T\textsuperscript{2}RAG by $+14.1$, and IRCoT by $+4.8$), 73.5 EM on 2Wiki, and 37.1 EM / 50.0 F1 on MuSiQue, ahead of HippoRAG2 (28.6 / 40.4), T\textsuperscript{2}RAG (26.6 / 37.7), and IRCoT (26.4 / 38.8). IRCoT, the strongest baseline on HotpotQA at 57.7 EM, illustrates the ceiling of graph-free iterative retrieval, which we exceed on every split. Feeding the same reader from dense retrieval alone (DPR-only) trails our full system by 14--42 EM per dataset, confirming that the KG-guided retrieval set, not the reader, drives the gains.

\emph{Ours (bounded)} tightens the token-reduction levers of \S\ref{sec:efficiency} until per-query cost matches T\textsuperscript{2}RAG's budget. Even so it stays ahead of T\textsuperscript{2}RAG on HotpotQA (57.2 vs.\ 48.4) and 2Wiki (67.7 vs.\ 55.2), i.e.\ both cheaper and more accurate. MuSiQue is the exception: its deep-hop questions are the least compressible, so EM under the bounded budget falls to 27.3.

\subsection{Retrieval Quality}

Table~\ref{tab:retrieval} contrasts our supporting-passage retrieval with dense retrieval alone. Dense retrieval attains decent recall but very low precision: some gold passages are present but drowned in distractors. Adding KG traversal converts this into a high-precision retrieval set at \emph{higher} recall on HotpotQA and 2Wiki, a roughly $2.5$--$3\times$ precision gain that directly explains the answer-quality gains in Table~\ref{tab:main}. On MuSiQue, recall reaches 68.0 at 55.6 precision, the strongest retrieval numbers of any system, though a large share of questions still land in partial recall, reflecting the difficulty of its deeper compositional chains (Section~\ref{sec:analysis}). LightRAG's corpus-constructed graph retrieves least effectively of all systems (43.2/67.9/34.8 recall), with the HotpotQA and MuSiQue shortfalls traced to its hybrid retriever returning empty results on roughly 40\% of short factoid questions (keyword extraction yields no low-level keywords, so no local context is retrieved).

\section{Analysis}
\label{sec:analysis}

\subsection{Component Ablations}

Table~\ref{tab:ablation} isolates the two retrieval mechanisms. Dense retrieval alone (DPR-only) gives high recall but low precision and weak answers; SPARQL-only traversal is precise but recall-limited to what KG title/$n$-gram matching anchors; interleaving the two (our full system) is strongest on every dataset. The two are complementary rather than redundant: on entity-named 2Wiki, SPARQL-only alone already reaches 69.6 EM, above every baseline, whereas HotpotQA and MuSiQue---whose entities are less often exact title matches---depend heavily on dense seeding ($+22$ and $+20$ EM from adding it).

\begin{table}[t]
  \centering
  \small
  \begin{tabular}{llccc}
    \toprule
    \textbf{Dataset} & \textbf{Variant} & \textbf{EM} & \textbf{AnsF1} & \textbf{R} \\
    \midrule
    \multirow{3}{*}{HotpotQA}
      & DPR-only     & 48.0 & 58.3 & 80.3 \\
      & SPARQL-only  & 40.2 & 49.8 & 54.0 \\
      & DPR+SPARQL   & \textbf{62.5} & \textbf{75.1} & \textbf{91.6} \\
    \midrule
    \multirow{3}{*}{2Wiki}
      & DPR-only     & 32.0 & 34.9 & 66.5 \\
      & SPARQL-only  & 69.6 & 76.0 & 85.7 \\
      & DPR+SPARQL   & \textbf{73.5} & \textbf{81.0} & \textbf{96.4} \\
    \midrule
    \multirow{3}{*}{MuSiQue}
      & DPR-only     & 13.0 & 18.8 & 57.0 \\
      & SPARQL-only  & 17.2 & 26.6 & 34.4 \\
      & DPR+SPARQL   & \textbf{37.1} & \textbf{50.0} & \textbf{68.0} \\
    \bottomrule
  \end{tabular}
  \caption{Component ablation at $n{=}1000$ (DPR-only at $n{=}100$). EM / answer-F1 / supporting-passage recall. DPR+SPARQL is our full system at the unbounded operating point. Dense seeding and KG traversal are complementary: SPARQL-only already suffices on entity-named 2Wiki (69.6 EM, beating every baseline), while HotpotQA and MuSiQue rely heavily on dense seeding ($+22$ and $+20$ EM), and combining the two is best on all three.}
  \label{tab:ablation}
\end{table}


\paragraph{Unbounded vs. bounded.} The unbounded setting pairs a lower-bar resolver with a full commit-oriented reader context; the bounded setting adds token-reduction levers to reach T\textsuperscript{2}RAG's per-query budget. The gap between them concentrates on questions with longer or partially-supported chains, where the tighter-budget resolver confirms too few passages and the reader is left to abstain. MuSiQue is the clearest case: the unbounded setting reaches 37.1 EM / 50.0 F1 (Table~\ref{tab:main}), against 27.3 / 36.1 for bounded. HotpotQA and 2Wiki, whose reasoning is shallower and whose retrieval is easier, are less sensitive, and the bounded setting stays within a few EM of unbounded on both. This is a knob along a cost--accuracy frontier, not a per-dataset choice: both settings use the same configuration across every benchmark.

\subsection{Parse vs.\ Execute}
\label{sec:parse_vs_execute}

To isolate the contribution of parsing, we replace it with direct execution: the raw SPARQL emitted by the plan agent is sent verbatim to the Wikidata endpoint, and its bindings are title-matched to corpus passages exactly as the parsed lookups are. Backbone, prompts, corpus, retriever, resolver, and reader are unchanged; property probing (\S\ref{sec:parse}) is disabled when executing, since it operates on parsed steps. We run the comparison at the bounded operating point, which is both where we compare against T\textsuperscript{2}RAG at matched budget and where the mechanism is exposed: with a single planning round and a five-passage dense pool, a failed KG query cannot be rescued by re-planning or by a large dense pool.

\begin{table}[t]
  \centering
  \small
  \setlength{\tabcolsep}{4pt}
  \begin{tabular}{lcccc}
    \toprule
    & \multicolumn{2}{c}{\textbf{EM}} & \multicolumn{2}{c}{\textbf{R}} \\
    \cmidrule(lr){2-3}\cmidrule(lr){4-5}
    \textbf{Dataset} & Parse & Execute & Parse & Execute \\
    \midrule
    HotpotQA & \textbf{57.2} & 56.9 & \textbf{79.8} & 77.9 \\
    2Wiki    & \textbf{67.7} & 56.3 & \textbf{87.2} & 78.3 \\
    MuSiQue  & \textbf{27.3} & 24.8 & \textbf{48.0} & 45.9 \\
    \bottomrule
  \end{tabular}
  \caption{Parsing vs.\ executing the generated SPARQL at the bounded operating point ($n{=}1000$; all else identical). \emph{Parse} is Ours (bounded) from Table~\ref{tab:main}.}
  \label{tab:parse_vs_execute}
\end{table}

Table~\ref{tab:parse_vs_execute} shows that executing the raw query is never better, and on 2Wiki it costs 11.4 EM and 8.9 recall points. This erases most of our bounded margin over T\textsuperscript{2}RAG (67.7 vs.\ 55.2 becomes 56.3 vs.\ 55.2). The size of the gap tracks how much each dataset depends on the KG rather than how hard it is: 2Wiki, whose questions are largely answerable from Wikidata relations, loses the most; on HotpotQA dense retrieval already does most of the work and the gap is negligible ($-0.3$); MuSiQue, whose entities Wikidata covers less well, falls in between ($-2.5$).

The outcomes of the executed queries explain why. Almost every raw query is well-formed: syntax errors account for only 0.3--2.4\% of queries and timeouts for at most 1.1\%. Yet 58--77\% return no bindings, because the query anchors on a hallucinated or mislinked identifier, matches a label string that does not exist verbatim (e.g., \texttt{rdfs:label "X"@en} with the wrong surface form), or chains constraints that no entity satisfies jointly. These are semantic failures rather than coding failures, so a stronger code-generation backbone would shrink the small syntax slice but not the dominant empty-result slice. Parsing sidesteps both: it keeps the model's choice of entities and relations, resolves label anchors through entity search rather than exact label match, drops placeholder identifiers, and replaces the joint pattern with independent anchored single-hop lookups that degrade gracefully when one hop fails. At the unbounded operating point the same comparison yields smaller but consistent gaps ($-1.6$ / $-3.6$ / $-0.9$ EM on HotpotQA / 2Wiki / MuSiQue), as re-planning across iterations and a larger dense pool recover many of the failed queries; there the mechanism improves accuracy but is not the only safeguard.

\subsection{Computational Efficiency: Tokens and LLM Calls}
\label{sec:efficiency}

We measure computational cost along two axes: the one-time \emph{corpus graph construction} paid before any query, and the recurring \emph{query-time} cost per question. All figures use exact token accounting on 2Wiki with the shared Qwen3-32B model; we run each method with a fresh LLM cache over the prebuilt corpus index so that every call is a real API call whose prompt/completion token usage is recorded. Query time and construction figures are for the 2Wiki corpus.

\paragraph{Construction cost.} GraphRAG, LightRAG, HippoRAG2, and T\textsuperscript{2}RAG all run an LLM over the entire corpus to extract entities, relations, or OpenIE propositions before any query can be answered, a cost that scales with corpus size and must be repaid whenever the corpus changes. HippoRAG2's OpenIE pass costs 8.1M tokens over 12{,}238 calls ($\approx$2 per passage); T\textsuperscript{2}RAG reuses the identical extraction, and LightRAG's entity/relation extraction plus community summarization costs an estimated 33.8M tokens (Table~\ref{tab:efficiency}, left). Our method incurs no such per-corpus pass because it reuses Wikidata, whose graph already exists. This is not free in an absolute sense, since Wikidata represents a large, ongoing human and institutional curation effort, but that cost is borne once, externally, and amortized across all users and tasks, rather than re-paid by each system on each corpus. The tradeoff is one of applicability rather than pure savings: where a curated KG covers the domain (as Wikidata does for the entity-centric questions in these benchmarks), our approach avoids the extraction step entirely; in a domain with no suitable existing KG, one would first have to be built or extracted, and the corpus-graph methods' construction cost would apply to us too. Table~\ref{tab:efficiency} therefore reports construction cost as the \emph{per-corpus LLM pass a method must run itself}, which is zero for ours on these benchmarks.

\paragraph{Query-time cost.} At query time (Table~\ref{tab:efficiency}, right) HippoRAG2 is cheapest at 1{,}437 tokens over a single reader call, with T\textsuperscript{2}RAG next at 3{,}218 tokens over 3.3 calls. Our full system uses 6{,}763 tokens over 4.6 calls: the DPR-seed, batch-resolution, and SPARQL-traversal loop. LightRAG is the most expensive per query (10{,}122 tokens).\footnote{LightRAG's figure includes an additional context-only pass we issue to recover retrieved passage titles for the recall metric; its native answer cost is lower. We report the measured upper bound.} The per-query cost of our system is not fixed: the token-reduction levers of this section trade traversal depth and prompt verbosity for tokens. \emph{Ours (bounded)} tightens them until per-query cost reaches 3{,}082 tokens over 3.6 calls, matching T\textsuperscript{2}RAG's budget, while, as Table~\ref{tab:main} shows, still exceeding T\textsuperscript{2}RAG in answer quality on HotpotQA and 2Wiki. At equal query-time budget our method is more accurate, and it can spend more to gain further accuracy when budget allows.

\begin{table}[t]
  \centering
  \small
  \setlength{\tabcolsep}{4pt}
  \begin{tabular}{lccrr}
    \toprule
    & \textbf{Corpus} & \textbf{Constr.} & \multicolumn{2}{c}{\textbf{Query-time}} \\
    \cmidrule(lr){4-5}
    \textbf{Method} & \textbf{graph} & \textbf{tokens} & \textbf{tok/q} & \textbf{calls/q} \\
    \midrule
    DPR-only    & No  & 0     & 1{,}228  & 1.0 \\
    HippoRAG2   & Yes & 8.1M  & 1{,}437  & 1.0 \\
    T\textsuperscript{2}RAG & Yes & 8.1M & 3{,}218 & 3.3 \\
    IRCoT       & No  & 0     & 3{,}390  & 2.4 \\
    LightRAG    & Yes & 33.8M & 10{,}122 & 3.0 \\
    \midrule
    Ours (bounded) & No & 0 & 3{,}082 & 3.6 \\
    Ours (unbounded) & No & 0 & 6{,}763 & 4.6 \\
    \bottomrule
  \end{tabular}
  \caption{Computational cost on 2Wiki (Qwen3-32B). \emph{Construction tokens} is the one-time LLM pass over the 6{,}119-passage corpus (entity/relation/OpenIE extraction) required before querying; it scales with corpus size and must be redone when the corpus changes. \emph{Query-time} tokens and calls are per question. T\textsuperscript{2}RAG reuses HippoRAG2's OpenIE extraction (same construction cost); LightRAG's construction is estimated by tokenizing its response cache. Ours and DPR-only run no per-corpus construction pass on these benchmarks, reusing a pre-existing KG / dense index; where no suitable KG exists, one would have to be built (\S\ref{sec:efficiency}). \emph{Ours (bounded)} applies the token-reduction levers to reach T\textsuperscript{2}RAG's query-time budget.}
  \label{tab:efficiency}
\end{table}

\subsection{Why We Beat the Baselines}

Part of our 2Wiki margin is a \emph{refusal} gap, which we quantify rather than leave implicit. Scoring every system's answers with a refusal detector (``not mentioned'', ``cannot be determined'', and similar), HippoRAG2 abstains on \textbf{34.3\%} of 2Wiki questions at $n{=}1000$, against \textbf{0.2\%} for our unbounded and \textbf{7.2\%} for our bounded setting. Since 2Wiki answers are short structured entities and dates, an abstention is a guaranteed miss, so a system that commits is advantaged independently of retrieval quality. This confound does not account for the whole gap: even restricted to the questions HippoRAG2 does answer, its exact match is 53.6, still well below our 73.5 (unbounded) and 67.7 (bounded) over all 1{,}000, so a margin of $+19.9$ (unbounded) and $+14.1$ (bounded) survives after neutralizing refusals. The remaining difference reflects genuinely better retrieval together with a lower abstention rate that is itself a product of the interleaved design: the resolver supplies confirmable evidence and the reader is prompted to commit. When retrieval succeeds (recall $=100\%$), our reader usually answers correctly; the residual losses on HotpotQA and 2Wiki are overwhelmingly reader-side rather than retrieval-side (75--84\% of their errors occur despite perfect recall; Appendix~\ref{app:errors}), so on those datasets answer generation, not retrieval, is the ceiling. MuSiQue inverts this---roughly two-thirds of its errors are retrieval-limited, and its deeper compositional chains are where the resolver's relevance bar and the reader's commit policy matter most: the unbounded setting lifts EM to 37.1 and F1 to 50.0, ahead of all three baselines.

\section{Conclusion}

We showed that a live external knowledge graph can be traversed effectively for multi-hop retrieval without building a corpus graph, by having an LLM plan traversal as SPARQL and then \emph{parsing rather than executing} it into reliable single-hop lookups. Interleaved with dense seeding and a batch query-refinement loop, this outperforms strong reproduced graph-based baselines on all three benchmarks under matched conditions, at lower LLM cost and, when a curated KG already covers the domain, without the per-corpus graph-construction pass those baselines require. A single configuration, applied identically to every dataset, wins throughout; on the hardest split, MuSiQue, it reaches 37.1 EM, ahead of all three baselines, and a token-matched bounded setting stays ahead of T\textsuperscript{2}RAG on HotpotQA and 2Wiki while spending roughly its per-query budget.

\section{Limitations}

Our approach depends on the availability and coverage of an external knowledge graph. We avoid a per-corpus construction pass by reusing Wikidata, but this shifts rather than eliminates the cost: Wikidata itself is the product of large-scale curation, and in a domain not covered by an existing KG a graph would first have to be built or extracted, at which point the construction cost of the corpus-graph baselines would apply to us as well. Even where a KG exists, entities absent from it cannot be traversed, which bounds recall on niche-entity questions. The method targets benchmarks whose passage titles align with KG entity labels; corpora without such alignment would require an entity-linking step we do not evaluate.

The approach also needs a named entity to anchor traversal on. A query with no such entity (e.g., ``who has won the most Grand Slams?'') reduces the contribution of the KG. Most information-seeking queries are entity-bearing, however: over 70\% of web-search queries contain a named entity \citep{kasturia2022query}, and Wikidata covers such entities broadly, so the only added cost is entity linking, with no per-corpus graph to build. On our benchmarks, roughly 7--11\% of questions per dataset contain no named entity (by NER); for these, dense retrieval provides robustness. For a specialized corpus that Wikidata does not cover (e.g., QA over financial reports), a graph would have to be constructed at a cost similar to T\textsuperscript{2}RAG's, removing our offline efficiency advantage, although we would expect the accuracy benefits of interleaving to carry over.

MuSiQue remains our hardest split even at the unbounded operating point (37.1 EM): its deepest questions require composing three or four hops, and the plan agent cannot always anchor a first hop whose start entity is described relative to a prior hop rather than named. All experiments use a single reader LLM (Qwen3-32B); we did not study sensitivity to reader choice.

\section{Ethical Considerations}
We used a smaller model on these academic benchmarks to show the validity of our approach and to reduce the impact on the environment. 
Throughout the development of this work, generative AI systems were employed for language refinement and code test case generation. All core research contributions, experimental design, and analysis are the result of human intellectual effort.

\bibliography{anthology}

\appendix

\section{Prompts}
\label{app:prompts}

We reproduce the prompts used by each agent verbatim. All are single-shot
instructions to Qwen3-32B at temperature $0$; \texttt{\{\dots\}} denotes runtime
substitution.

\subsection{Plan Agent}
The plan agent is prompted with a system message (task framing + a retrieved
Wikidata property schema + one worked example) followed by a per-iteration user
message carrying the running state. The system message opens:

\begin{quote}\small\ttfamily
You are an expert in Wikidata and SPARQL. Your task: generate SPARQL queries that
express what graph traversal paths to explore\dots{} The SPARQL you generate will
NOT be executed directly. Instead, it will be parsed to extract (entity, property,
direction) traversal tuples. So: focus on expressing the RIGHT entities and
properties; use \mbox{rdfs:label} "Name"@en to anchor to named entities; use wdt:PXX
for properties; keep queries simple and clear --- they're a communication format,
not executable code.
\end{quote}

\noindent An agentic suffix casts it as a fact-discovery loop:

\begin{quote}\small\ttfamily
You receive: the question; facts already resolved; what information is still
needed; available properties on known entities (from probing); dead ends
(properties that returned nothing --- DO NOT repeat these). CRITICAL RULES: ALWAYS
anchor queries to SPECIFIC named entities. NEVER query abstract classes. Focus on
finding the MISSING information. Use CONFIRMED properties from the probed list. DO
NOT repeat dead ends.
\end{quote}

\noindent The per-iteration user message fills \texttt{\{question\}},
\texttt{\{resolved\_info\}}, \texttt{\{still\_needed\}},
\texttt{\{probed\_properties\}}, and \texttt{\{dead\_ends\}}.

\subsection{Resolution Agent}
The resolver judges a whole candidate passage set in one call. The
precision-oriented (bounded) prompt instructs:

\begin{quote}\small\ttfamily
You are given several candidate passages at once. For EACH passage, decide whether
it contains facts DIRECTLY on the reasoning path to the answer\dots{} CRITICAL ---
Be STRICT about relevance: ONLY mark a passage relevant if it contains facts on the
direct reasoning path. A passage about a different person/event/place with a
similar name is NOT relevant. Generic/definitional passages are NOT relevant.
\end{quote}

\noindent It emits a fixed format (\texttt{RELEVANT}, \texttt{FACTS},
\texttt{STILL\_NEEDED}, \texttt{REFINED\_QUERY}, \texttt{STOP}, \texttt{ANSWER}).
The recall-oriented (unbounded) prompt keeps the same output format but replaces
the strictness block with an inclusive one:

\begin{quote}\small\ttfamily
This is a MULTI-HOP question\dots{} BE INCLUSIVE about relevance: mark a passage
relevant if it helps with the FINAL answer OR with ANY intermediate hop\dots{} When
in doubt, INCLUDE the passage. Missing a bridge passage breaks the chain. Only
exclude passages clearly about a DIFFERENT entity that merely shares a similar name.
\end{quote}

\subsection{Reader}
The default reader reasons step-by-step and emits a short span
(``\texttt{Thought:}\dots{}\texttt{Answer:}''), with three few-shot examples
(a bridge, a comparison, and a yes/no question). The unbounded setting swaps in a
\emph{commit} variant that forbids abstention:

\begin{quote}\small\ttfamily
ALWAYS commit to a concrete answer. Even if the passages are incomplete or the
answer requires combining facts across passages and your own knowledge, give your
single best guess. NEVER answer "unknown", "not stated", "insufficient
information", or "cannot be determined". These are always wrong.
\end{quote}

\section{SPARQL Parser}
\label{app:parser}

The parser converts a generated SPARQL query into an ordered list of traversal
steps $(e_i, r_i, d_i)$ without ever executing it. It applies the following rules
to each query:

\begin{enumerate}
\item \textbf{Strip non-traversal blocks.} \texttt{SERVICE wikibase:label},
\texttt{FILTER}, and \texttt{FILTER NOT EXISTS} blocks are removed --- they affect
formatting/execution but carry no traversal edge.
\item \textbf{Label anchors.} \texttt{?x rdfs:label "Name"@en} binds variable
\texttt{?x} to a named entity; the name is later resolved to a QID via the
entity-search API.
\item \textbf{Direct QID anchors.} \texttt{wd:Q42 wdt:P69 ?u} yields an
\emph{outgoing} step from \texttt{Q42} along \texttt{P69}.
\item \textbf{Incoming patterns.} \texttt{?w wdt:P57 wd:Q25191} yields an
\emph{incoming} step into \texttt{Q25191} along \texttt{P57}.
\item \textbf{Discard type constraints.} Triples with \texttt{P31} (instance of) or
\texttt{P279} (subclass of) are dropped; they constrain types rather than traverse.
\item \textbf{Drop invented placeholders.} Round-number QIDs
(e.g.\ \texttt{wd:Q12345}) the model emits when it does not know a real ID carry no
label hint and are skipped.
\item \textbf{Chain dependent hops.} When one hop's output variable feeds the next
(\texttt{?v1 wdt:P1 ?v2 . ?v2 wdt:P2 ?v3}), the second step is anchored to a
\texttt{\$RESULT} reference so it executes against the first hop's resolved entity.
\end{enumerate}

\noindent Each surviving step runs as an anchored single-hop query
(\texttt{wd:QID wdt:$r_i$ ?v} or its reverse), which binds a concrete entity and a
single triple pattern and therefore cannot time out or raise a
query-complexity error.

\paragraph{Worked example.} For ``\emph{Who directed the film that Leonardo
DiCaprio starred in about dreams?}'' the plan agent emits:

\begin{quote}\small\ttfamily
\# Find directors of films starring Leonardo DiCaprio\\
SELECT ?director WHERE \{\\
\phantom{xx}?actor rdfs:label "Leonardo DiCaprio"@en .\\
\phantom{xx}FILTER NOT EXISTS \{ ?actor wdt:P31 wd:Q4167410 \}\\
\phantom{xx}?film wdt:P161 ?actor .\\
\phantom{xx}?film wdt:P57 ?director .\\
\phantom{xx}SERVICE wikibase:label \{ \dots{} \}\\
\} LIMIT 50
\end{quote}

\noindent The parser drops the \texttt{FILTER} and \texttt{SERVICE} lines, binds
\texttt{?actor} to ``Leonardo DiCaprio,'' and produces two steps:
(\texttt{Leonardo DiCaprio}, \texttt{P161}, incoming) --- films with DiCaprio in
the cast --- then (\texttt{\$RESULT}, \texttt{P57}, outgoing) --- their directors.
Neither is executed as written SPARQL; both become anchored single-hop lookups.

\section{MuSiQue Corpus De-duplication}
\label{app:dedup}

The MuSiQue retrieval corpus released by HippoRAG contains 11,656 passages stored
as \texttt{\{title, text\}} pairs with no passage identifier. 647 titles are
duplicated across 2,465 passages (homonyms, e.g.\ 28 distinct passages titled
``New York City''). A naive corpus map keyed by title collapses these duplicates
and silently drops 1,818 passages, so only 9,838 ever enter the index or reach the
reader; supporting-passage recall then becomes title-level and optimistic. We key
the corpus, index, and gold labels by a synthetic passage id
(\texttt{title::hash(text)}) so that all 11,656 passages are indexed and recall is
measured at the passage level. This matches how the graph-based baselines
(HippoRAG2, T\textsuperscript{2}RAG) already ingest the corpus, which chunk by
content rather than title; the change corrects our own ingestion rather than
altering the shared benchmark. Beyond honest measurement, it materially helps
answering: feeding the reader the correct homonym passage rather than a collapsed
survivor raises MuSiQue EM from 29.4 to 37.1 (Appendix~\ref{app:anecdotes} gives a
concrete case). HotpotQA and 2Wiki corpora have unique titles and are unaffected.

\section{Error Decomposition}
\label{app:errors}

To locate where each dataset's remaining errors come from, we bucket every wrong
answer ($\text{EM}=0$) by supporting-passage recall: \emph{reader failures} (recall
$=100\%$ --- the gold passages were retrieved but the reader did not produce the
exact answer), \emph{partial retrieval} ($0<\text{recall}<100\%$), and \emph{zero
retrieval}. Table~\ref{tab:errors} shows a sharp split by dataset.

\begin{table}[t]
  \centering
  \small
  \begin{tabular}{lcccc}
    \toprule
    \textbf{Dataset} & \textbf{Wrong} & \textbf{Reader} & \textbf{Partial} & \textbf{Zero} \\
    \midrule
    HotpotQA & 375 & 283 (75\%) & 74 (19\%) & 18 (4\%) \\
    2Wiki    & 265 & 225 (84\%) & 39 (15\%) & 1 (0\%) \\
    MuSiQue  & 629 & 198 (31\%) & 375 (60\%) & 56 (9\%) \\
    \bottomrule
  \end{tabular}
  \caption{Wrong answers ($n{=}1000$, unbounded) bucketed by supporting-passage
  recall. \emph{Reader}: perfect recall but wrong answer; \emph{Partial}/\emph{Zero}:
  retrieval-limited. HotpotQA and 2Wiki errors are overwhelmingly reader-side (the
  gold passages are retrieved), whereas MuSiQue errors are dominated by
  retrieval---its deeper compositional chains break before the evidence is assembled.}
  \label{tab:errors}
\end{table}

On HotpotQA and 2Wiki, 75--84\% of errors occur despite \emph{perfect} retrieval:
the coupled retrieval set already contains the gold passages, and the residual loss
is the reader emitting a near-miss span rather than the exact gold string. Retrieval
is effectively solved on these two, and answer generation is the ceiling. MuSiQue
inverts this: only 31\% of its errors are reader-side, while 69\% are
retrieval-limited (partial or zero recall). Of the 629 MuSiQue errors, 247 produced
\emph{no} SPARQL facts at all---the plan agent never anchored a traversal, the
described-entity failure mode of Appendix~\ref{app:anecdotes}---confirming that
MuSiQue's bottleneck is assembling the reasoning chain, not reading it.

\section{Qualitative Examples}
\label{app:anecdotes}

\paragraph{A full trace (and why de-duplication matters).} Consider the MuSiQue
question ``\emph{How many times did the plague occur in the city where the painter
of The Bacchanal of the Andrians died?}'' (gold answer: 22). Dense seeding retrieves the
painting's passage, from which the resolver extracts that Titian
(\texttt{Q47551}) painted it. The plan agent writes SPARQL anchoring Titian and
traversing \texttt{P20} (place of death); the parser turns this into the single-hop
lookup \texttt{Q47551 ${\to}$ P20 ${\to}$ Venice}. The resolver folds ``died in
Venice'' into a refined query, which re-seeds dense retrieval and surfaces the
plague-in-Venice passage. On the collapsed corpus the reader receives the wrong
``Black Death'' homonym text and answers ``Once'' (EM $=0$); on the
de-duplicated corpus (Appendix~\ref{app:dedup}) it receives the passage stating
``plague occurred in Venice 22 times between 1361 and 1528'' and answers ``22''
(EM $=1$). The same traversal succeeds either way --- what changes is whether the
reader is handed the correct passage text.

\paragraph{Property probing corrects relation choice.} A common plan-agent error is
selecting a plausible but wrong property, e.g.\ \texttt{P570} (date of death) when
the question needs \texttt{P20} (place of death). After each traversal we probe the
discovered entity for its available properties and feed the list back to the plan
agent, which then re-issues the hop with the correct relation --- turning a dead end
into a resolved fact without a wasted iteration.

\paragraph{Failure 1: an entity described, not named.} For ``\emph{What is the
birthplace of the person after whom S\~{a}o Jos\'{e} dos Campos was named?}'' (gold:
Nazareth), the first hop's start entity is \emph{described} (``the person after whom
$X$ was named'') rather than named. The plan agent has no label to anchor on, so it
emits no traversable SPARQL: the run ends at iteration~0 with zero SPARQL facts and
no confirmed passages, and the reader guesses ``Bethlehem.'' This is the dominant
MuSiQue failure mode --- the plan agent cannot anchor a hop whose start entity is
defined only relative to a later hop.

\paragraph{Failure 2: dense seeding anchors the wrong entity.} For ``\emph{Who is
the spouse of the actor that plays Jarvis in the Avengers: Age of Ultron?}'' (gold:
Jennifer Connelly, spouse of Paul Bettany), dense retrieval surfaces a
Samuel~L.~Jackson / Nick~Fury passage instead of Paul~Bettany / Jarvis. The resolver
confirms the wrong actor, the traversal chain is anchored on the wrong entity, and
the answer collapses to ``Unknown.'' Here the failure is upstream of traversal: the
seed itself is wrong, so no amount of correct KG traversal recovers the gold path.

\end{document}